\def\year{2020}\relax
\documentclass[letterpaper]{article} 
\usepackage{aaai20}  
\usepackage{times}  
\usepackage{helvet} 
\usepackage{courier}  
\usepackage[hyphens]{url}  
\usepackage{graphicx} 
\usepackage{amsfonts}
\usepackage{subcaption}
\usepackage{booktabs}
\usepackage{array}
\usepackage{graphicx}  
\newcommand{\citet}[1]{\citeauthor{#1} \shortcite{#1}}
\newcommand{\citep}{\cite}

\usepackage{color,soul}

\title{On the use of Pairwise Distance Learning\hspace{10cm} for Brain Signal Classification with Limited Observations}
\author{ \Large \textbf{David Calhas,\textsuperscript{\rm 1,3} Enrique Romero,\textsuperscript{\rm 2,4} Rui Henriques\textsuperscript{\rm 1,5}}\\ 
\textsuperscript{\rm 1}Instituto Superior Tecnico, Lisbon, Portugal\\ 
\textsuperscript{\rm 2}Universitat Politecnica de Catalunya, Barcelona, Spain\\
\textsuperscript{\rm 3} david.calhas@tecnico.ulisboa.pt, \textsuperscript{\rm 4} eromero@cs.upc.edu, \textsuperscript{\rm 5} rmch@tecnico.ulisboa.pt
}
 
\begin{document}

\maketitle

\begin{abstract}
The increasing access to brain signal data using electroencephalography creates new opportunities to study electrophysiological brain activity and perform ambulatory diagnoses of neurological disorders. This work proposes a pairwise distance learning approach for Schizophrenia classification relying on the spectral properties of the signal. Given the limited number of observations (i.e. the case and/or control individuals) in clinical trials, we propose a Siamese neural network architecture to learn a discriminative feature space from pairwise combinations of observations per channel. In this way, the multivariate order of the signal is used as a form of data augmentation, further supporting the network generalization ability. Convolutional layers with parameters learned under a cosine contrastive loss are proposed to adequately explore spectral images derived from the brain signal. Results on a case-control population show that the features extracted using the proposed neural network lead to an improved Schizophrenia diagnosis (+10pp in accuracy and sensitivity) against baselines, suggesting the existence of non-trivial electrophysiological brain patterns able to capture discriminative neuroplasticity profiles among individuals.
\end{abstract}

\section{Introduction}\label{section:introduction}

The recording of increasingly affordable and precise electroencephalographic (EEG) data is creating unprecedented opportunities to understand brain activity, aid personalized prognostics, 
and promote health through wearable biofeedback systems \cite{nan2012individual}.
Electroencephalography is non-invasive, safe, inexpensive, and shows rich temporal content; in contrast with other brain imaging modalities, such as magnetic resonances, entailing higher costs, risks and restrictions on the periodicity of recordings \cite{fowle2000uses}. EEG monitoring is widely used to assess psychiatric disorders, and has shown to be a valuable source to study Schizophrenia, a disorder affecting about 1\% of the world population, largely susceptible to misdiagnoses \cite{owen201686}.

Despite the inherent advantages of monitoring electrophysiological brain activity, its use for diagnosing neuronal diseases is still capped by the limited size of case-control populations \cite{litjens2017survey}, as well as by the intrinsic difficulties of mining brain signals. 
Brain signal data is high-dimensional, multivariate, susceptible to noise/artefacts, rich in temporal-spatial-spectral content, and highly-variable between individuals \cite{lopes_2013}. 

This work proposes a dedicated class of neural networks to extract discriminative features of Schizophrenia from electrophysiological brain data. The proposed approach combines principles from pairwise distance learning and spectral imaging in order to address the aforementioned challenges, enabling superior diagnostics. 
Accordingly, the proposed approach offers seven major contributions:
\begin{enumerate}
\item Ability to learn from small datasets by taking advantage of Siamese Network layering, inherently prepared to work in augmented data spaces mapped from a limited number of observations \cite{eeg_dataset}. The features produced by these networks have proven to be useful to perform classification as they rely on either the homologous or discriminative properties of observation-pairs in a pairwise distance domain \cite{koch2015siamese};
\item Ability to deal with the rich and complex spectral and temporal content of EEG data by processing the signal into spectral images with a fine frequency and temporal resolution per electrode, and by subsequently reshaping the Siamese network architecture with adequate convolutional operations;
\item Robustness to noise and wave-instability by assessing distances on the spectral content under a cosine-loss. Gathered evidence shows less susceptibility to artefacts and the inherent variability of electrophysiological potentials associated with continuously changing overlapping electrical fields produced by localized neurons \cite{lopes_2013};
\item Ability to deal with the multivariate nature of the signal (rich spatial content) by capturing interdependencies between channels as their content is simultaneously used to shape the weights of shared connections in the network;
\item Ability to handle the extremely-high dimensional nature of the gathered spectral content from brain signals (high-resolution spectral image per electrode) under L1 regularization;
\item Applicability of the proposed EEG-based diagnostics to alternative populations or diseases, evidenced by the: i) placed Bayesian optimization step \cite{snoek2012practical} for hyperparameter tuning and fixing feature numerosity; ii) fully-automated nature of the approach once signals are recorded; and iii) generalization ability of the learning process on validation data.
\end{enumerate}

In contrast with the traditional stance to neural information processing systems, this manuscript explores whether we can go deep on highly-dimensional spatiotemporal data in the presence of a very limited number of data observations. 
This stance is much needed in healthcare given the limited size of trials (cohort studies), often driven by disease rarity, capped size of control population, trial eligibility requirements, or the facultative nature of EEG assessments. 
Results confirm this possibility: +10pp in the accuracy and sensitivity of Schizophrenia diagnostics. 

The features extracted from the proposed spectral and pairwise distance space further suggest the presence of discriminative elecrophysiological patterns linked to neuroplasticity aspects of the individuals. This observation is in accordance with findings from previous studies that established statistically significant relationships between variations in the frequency band spectrum and neuroplasticity conditions \cite{bhandari2016review,liu2015plasticity}.






The manuscript is organized as follows. After formalizing the problem, Section \ref{section:related_work} surveys existing contributions on the diagnosis of individuals from brain signal data. Section \ref{section:approach} describes the proposed solution. Section \ref{section:results} shows extended evidence of its relevance for diagnosing Schizophrenia. Finally, concluding remarks are drawn in Section \ref{section:conclusion}. 

\subsection{Problem Description}


\noindent\textbf{Problem.} A EEG recording or brain signal observation is a multivariate time series $X$=$\{x^j_t\mid j\in\{1..M\}, t\in\{1..T\}\}$, where $x^j_t$ is a measure of the electrophysiological activity in scalp channel $j$ and instant $t$, $T$ is the number of time points, and $M$ is the multivariate order (number of channels). Given a brain signal dataset $\{(X_i,c_i)\mid i=1..N\}$, with $N$ EEG recordings $X_i$ annotated with a label $c_i\in\Sigma$, our task is to identify a discriminative feature space to classify (unlabeled) observations. Specifically, we are interested in classifying Schizophrenia given case-control populations. 
\vskip 0.3cm

\noindent\textbf{Background.} The electrophysiological signal produced by a specific channel $k$ in the cerebral cortex is a univariate time series that can be decomposed into a frequency time series using a discrete Fourier transform. The analysis of the frequency domain of a signal, generally referred as spectral analysis, determines the predominant waves monitored at a certain location. A short-time discrete Fourier transform can be alternatively applied along a sliding window of the raw signal to capture potentially relevant changes on the spectral activity of the brain throughout the EEG recording. The spectral content produced by this time-varying form of spectral analysis is here informally referred as a \textit{spectral image} since it measures brain activity along two contiguous axes: time and frequency. 

\section{Related Work}\label{section:related_work}

\begin{table*}[ht]
    \centering
    \footnotesize
    \caption{\small Schizophrenia EEG datasets.}
    
    \begin{tabular}{m{5cm} m{3cm} m{4cm} m{1.5cm}}
        \toprule
        Dataset & Healthy Controls & Schizohprenic Individuals & Access \\
        \midrule
        \citet{dvey2017connectivity} and \citet{dvey2015schizophrenia} & 20 & 20 & Private\\
        \citet{sabeti2009entropy} & 25 & 25 & Private\\
        \citet{eeg_dataset} & \textbf{39} & \textbf{45} & \textbf{Public}\\
        \bottomrule
    \end{tabular}
    \label{table:datasets}
\end{table*}

\subsection{EEG Classification}

EEGNet \cite{eegnet}, EEGNet-SSVEP \cite{eegnet}, DeepConvNet \cite{schirrmeister2017deep} and ShallowConvNet \cite{schirrmeister2017deep} are considered state-of-the-art EEG classification built models that make use of convolutional operations directly on the raw EEG data. These convolutions are placed along time and channels. Approaches like these rely on the properties of its models to extract discriminative features from EEG signals. These models are validated with a total of 4 datasets, all of which are based on EEG task or stimuli based recording sessions. One can see directly that these networks learn event related potentials from the EEG signal, which makes the EEG recording session dependable of a task environment. In contrast, we aim at extracting neuroplasticity related features from the EEG signal, as the dataset used is based on a resting state recording session. In Section \ref{section:results}, these models are shown to perform bad on resting state EEG data.

\subsection{EEG on Schizophrenia}



\citet{dvey2015schizophrenia} claim mostly changes in functional connectivity are seen in patients with Schizophrenia, as well as differences in theta-frequency activity. A classification approach was applied on 1-minute signals recorded by a single electrode. 
The developed system consists of four stages: performing several preprocessing tasks and breaking the raw signals into relevant intervals; transformation of the EEG signal into a time/frequency representation via the Stockwell transformation; feature extraction from the time/frequency representation; and discrimination of specific time frames following a given set of stimuli between the time/frequency matrix representations of the healthy subjects and the schizophrenia patients.
Despite promising results, the approach requires the performance of cognitive tasks by the individuals under assessment throughout the recording. 
\citet{dvey2017connectivity} introduced another way of looking at the EEG signal using connectivity maps derived from the brain activity. 
In order to build these maps, a similarity function needs to be chosen, so one can check which nodes are more similar to which ones. 
Results showed that the degradation of connectivity is being accelerated within schizophrenia individuals. And that information relay changes in an abnormal manner primarily in the prefrontal area. 
This gives a good insight on how connectivity maps can be applied to discriminate schizophrenia. And most important, that one should take into account that a change in a certain region can influence other regions in the brain.

\citet{sabeti2009entropy} introduced another approach to classify Schizophrenia based on entropy and complexity measures of the EEG signal.
The features extracted from the signal were: Shannon entropy, spectral entropy, approximate entropy, Lempel-Ziv complexity and Higuchi fractal dimension. Genetic programming was used for feature selection. With these features, Adaptative boost (Adaboost) and Linear Discriminant Analysis (LDA) classifiers were validated, showing performance improvements against peer approaches. 
The recordings were done with eyes open, a setting easily biased by environmental effects. 

Notable examples of connectionist and spectral approaches were introduced 
to discriminate and characterize Schizophrenia. Nevertheless, there is still a research gap on how to simultaneously explore the rich spectral, temporal and spatial nature of brain signals to perform classification. 
In spite of the indisputable role of neural network learning for the analysis of complex spatiotemporal signal data, its role for EEG-based diagnostics of psychiatric disorders remains largely unexplored due to the absence of large cohorts and the inherent stochastic complexities associated with electrophysiological data. 

\subsection{Siamese Neural Network}

First introduced by \citet{bromley1994signature} as a novel model used in the task of signature classification whose aim  was to distinguish signature forgeries from the real ones, Siamese Neural Networks (SNN) are deep learning architectures with two sub-networks that consist on the same instance, hence being called "siamese networks". This architecture receives as input a pair of samples. Subsequently, the outputs of the pairs used as input to these "siamese networks" are joined in a distance function. The proposed distance function between the output of the SNNs is the cosine similarity (for signatures from the same person the output should be $1$, and $-1$ for forged ones). This model had outstanding results at the time, detecting $80.0\%$ of the forged signatures and $95.5\%$ of the genuine signatures. More recently, \citet{koch2015siamese} successfully used a SNN Architecture for One Shot Learning (meaning the model only sees each class once in an epoch). This approach reached $92.8\%$ accuracy in the test set. These results were achieved through a Siamese Convolutional Architecture. Once this kind of network is trained, its learned representations via a supervised metric-based approach with SNNs are useful to perform tasks like classification, relying on the discriminative properties of these features.

\section{Our Approach}\label{section:approach}

The proposed architecture is inspired by the architecture formerly introduced by \citet{koch2015siamese}. An advantage of this type of architecture is the ability to augment the original dataset from an instance-based data space to a pair-based one. Our approach has two main steps: 1) feature extraction; and 2) classification. In step 1, the internal representations obtained from the SNN architecture model are extracted after training. In step 2, a classification task is performed using these extracted features. Previous to both steps, we perform hyperparameter optimization for every model using Bayesian Optimization (BO) \cite{snoek2012practical}.

\subsection{Dataset Description}

Approaches based on induced stimuli or task performance, followed by the analysis of event related potentials, are not considered in this work. 
Instead, a resting state setup is consider to monitor the underlying brain patterning at the brain cortex, 
independently of the surrounding environment/undertaken task. Subsequently, this avoids any additional interference on the EEG signal recorded. 
\citet{howells2018electroencephalographic} findings support the use of this setup, claiming that differences on the spectral activity -- such as higher delta and a lower alpha synchronization in psychotic disorders -- can be optimally detected in resting state protocols with both open and closed eyes.

Table \ref{table:datasets} shows the content of EEG datasets containing healthy control individuals and schizophrenic individuals. \citet{dvey2017connectivity}, \citet{dvey2015schizophrenia} and \citet{sabeti2009entropy} works were introduced and discussed in Section \ref{section:related_work}. Unfortunately, besides their low significance and variance, due to the lack of observations, the datasets used are not made publicly available. Nonetheless, \citeauthor{eeg_dataset} gathered a total of 84 individuals, of which 45 were schizophrenic and 39 were regarded as healthy controls. 

The population in \cite{eeg_dataset}, consists of adolescents who had been screened by a psychiatrist and got either a positive or negative diagnostic for the schizophrenia neuropathology. EEG recordings were sampled at $128$ Hz with $1$ minute duration. Individuals were set in a resting state with eyes closed. In accordance with the 10-20 system of electrode placement, the topographical positions of the placed EEG channels are: F7, F3, F4, F8, T3, C3, Cz, C4, T4, T5, P3, Pz, P4, T6, O1, O2.


\subsection{Siamese Neural Network Architecture}

The SNN architecture contains two sub networks that correspond to the same instance (twin networks). Both of these twin networks are referred to as the Base Network (BN). The input and output of the BN are an example and a feature vector, respectively. The output feature vector corresponds to the features extracted in the aforementioned step 1.

\begin{figure*}[ht]
  \centering
  \includegraphics[width=0.8\linewidth]{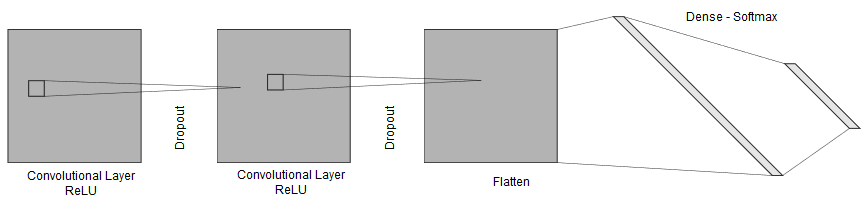}
  \caption{Base network from the SNN.}
  \label{fig:base_network}
\end{figure*}

In our case, the BN receives as input a Discrete Short-Time Fourier Transform (DSTFT) representation of the EEG signal, that is extracted from the 1 minute recording of a channel of an individual. The DSTFT is taken with 2 seconds length windows in order to capture frequencies as low as $0.5$Hz, corresponding to the delta wave frequencies (\citet{howells2018electroencephalographic} points out that frequencies lower than $2$Hz are relevant to differentiate Schizophrenia). This image is processed through two convolutional layers, followed by a fully connected layer. The activation function used in the convolutional layers is the rectified linear function \cite{hahnloser2000digital}, while the fully connected layer uses the softmax activation function, normalizing the domain of the feature representations, $\vec{f} \in \mathbb{R}^q, i \in [1, q]: f_i \in [0, 1]$.

\begin{figure}[ht]
  \centering
  \includegraphics[width=0.45\textwidth]{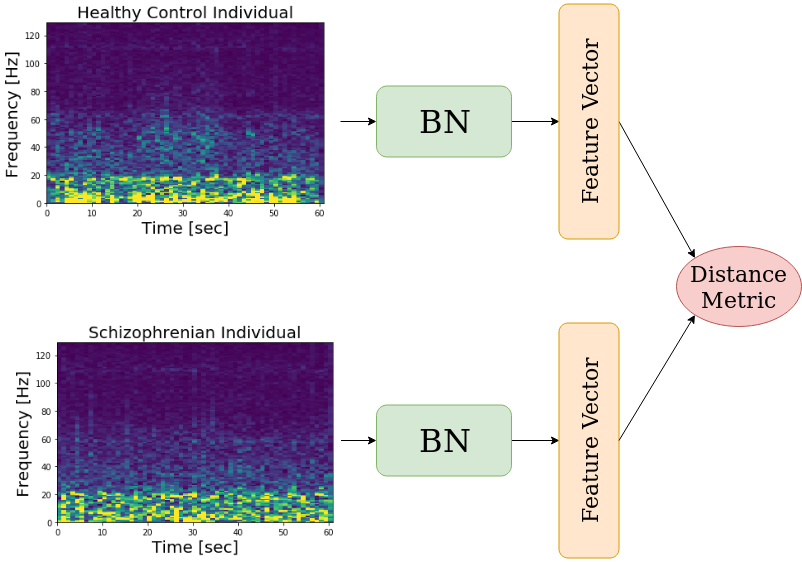}
  \caption{SNN architecture.}
  \label{fig:snn_architecture}
\end{figure}

Once the BN network (Fig. \ref{fig:base_network}) is built, a replication of it is made, producing its twin and sharing their weights. The SNN layout is achieved joining these twins and computing a distance metric between their outputs, as shown in Fig. \ref{fig:snn_architecture}. In our case, the inputs to the SNN are pairs of DSTFT representations and the outputs are the computed distance between the representations obtained by the BN.

The SNN tries to solve what is known as a neighbor separation problem, consisting on the separation of instances in a dataset that contains different classes. In our case we have two classes: schizophrenic and healthy control individuals. In this neighbor separation problem, pairs of individuals of the same class (schizophrenic with schizophrenic or healthy with healthy) are called neighbors and pairs of individuals of different classes (schizophrenic with healthy) are called non-neighbors. The network learns a transformation with the objective of assigning small distance to neighbors and large distance to non-neighbors.

With the previously described architecture, the neighbor separation problem can be posed as a minimization problem of a certain loss function that depends on such distance. In \cite{contrastiveLoss}, the Contrastive Loss function is introduced to that end, defined as:

\begin{equation}\label{equation:contrastive_loss}
L(W,Y,X_1,X_2) = Y{D_W}^2 + (1-Y) \ {max(0, m - D_W)^2}    
\end{equation}

\noindent where $(X_1,X_2$) is the input pair, $Y = 1$ if $X_1$ and $X_2$ are neighbors and $0$ otherwise, $D_W$ the distance between the predicted values of $X_1$ and $X_2$, and $m$ is the margin value of separation. Minimization of the Contrastive Loss function leads to a scenario where neighbors are pulled together and non-neighbors are pushed apart, according to a certain distance metric. The margin value is sensitive. High values of $m$ increase the separation between non-neighbors (pairs of different class), impacting positively the accuracy although making the training slower. In contrast, low values of $m$ may cause the model not to learn the desired behavior.

The distance metric considered in our case is the cosine distance. This metric was chosen with the belief that a pattern based metric (cosine) would perform better than a magnitude based one (euclidean), in order to shed light on how the schizophrenia pathology expresses itself through the EEG.

Besides the type of layers and the distance metric, the following techniques are integrated in the model: $L1$ regularization and Dropout layers. The $L1$ regularization is useful because it helps remove features that are not useful for the task. Dropout layers are introduced to improve generalization. Regularization is applied at the kernel of all layers. The Dropout probability used is $0.5$, as suggested by \citeauthor{srivastava14a}, and is applied after each convolutional layer. Adam \cite{adam} is used to optimize the network during the training session.

\subsubsection{Hyperparameter Tuning}

The number of layers, as well as their type, are fixed. The rest of hyperparameters (regularization factor, margin value, learning rate, kernel size and output dimension of the BN) are susceptible to optimization. As previously mentioned, we apply BO to that end. BO is set to run with a maximum of $50$ acquisitions and starts with $5$ iterations to perform an initial exploration. In each iteration and acquisition, a $K$-fold Cross Validation with $K=5$ is done with the training set of a Leave-One-Out Cross Validation (LOOCV) partition. The combination of hyperparameters that has the best average validation accuracy across the $5$-folds is chosen to perform the feature extraction. Each of the hyperparameters are assigned the following value domains to explore: regularization factor $ \in [10^{-3}, 10^{-1}]$, margin value $ \in [1.0, 2.0]$ , learning rate $ \in [10^{-6}, 10^{-3}]$, kernel size $t \times f$ with $t = f = \{3, 4, 5, 6, 7, 8, 9, 10, 11, 12\}$ (the same kernel size is used for both convolutional layers) and final output dimension $\in \{2, 4, 6, 8, 10, 12, 14\}$. The BO surrogate model is a standard Gaussian Process. Expected Improvement is used as an acquisition function and the Limited-Memory Broyden–Fletcher–Goldfarb–Shanno algorithm as the acquisition optimizer.

The DSTFT magnitudes are normalized, under the hypothesis that there exists a threshold from which there is no additional information to identify the schizophrenia pathology. With this, the values are normalized by an upper value, $U$. Values of $f$ smaller than $U$ are divided by $U$ and magnitudes bigger than $U$ are set to $1.0$. This allows every magnitude of the frequencies to be within the interval $[0, 1]$ after the normalization is performed. We take advantage of the BO exploration to obtain $U$, by introducing it in the same optimization process made for the SNN hyperparameters. The domain assigned to be explored for $U$ is $[100.0, 500.0]$.

\subsubsection{Pairwise Dataset Structure}

We want the network to learn a valid transformation that generalizes to all channels. To that end, the pairs are set such that only equal channels are paired. Pairs of different channels are not considered, since we see different channels as correlated spaces with different properties. Fortunately, the SNN is capable of learning different spaces/classes, as shown in \cite{koch2015siamese}, where the proposed system is able to learn a similar setup. This pairwise schema can be seen as a data augmentation technique being performed with the addition of noise to the dataset. This noise is present by mixing all the channels with the aforementioned restrictions in order to be coherent. No other data augmentation scheme, such as image transformations (scaling, rotations), is applied.

From our original EEG dataset, ${X_1, ..., X_N}$ spectral images are derived with $N=84$ examples, and a pairwise dataset $P$ is built. Formally, $P = {P_1, ..., P_{O}}$ with $O=c\ \mathrm{C}_{2}^{N}=M\ \mathrm{C}_{2}^{84}=55776$, where $M=16$ is the number of EEG channels. The space complexity of the pair dataset is $\mathcal{O}(c\ \mathrm{C}_{2}^{N})$. The SNN training session is done with a batch size multiple of the number of channels. In particular, we use $B=16*c$. Therefore, there are $16$ pairs of individuals in each batch and each pair of individuals has $c=16$ channel pairs. This scheme can only be applied in small datasets, since the model does not scale well in terms of space complexity, but our goal is precisely to tackle small datasets by the creation of a whole new optimization space where the variability contained in the data can be exploited in a different way.

\subsection{Validation}\label{subsection:validation}

After the SNN has been tuned and trained (in a $20$ epochs session), the outputs of the BN for every example were the result of our feature extraction process. With these features, the following classifiers were trained to identify schizophrenia: Support Vector Machines (SVM), Random Forest (RF), XGBoost (XGB), Naive Bayes (NB) and k-Nearest Neighbors (kNN). This process was performed with a LOOCV, where each fold consists on one subject ($16$ channels/instances). For each of these classifiers, BO hyperparameter tuning is also performed, setup with a maximum of $10$ acquisitions and $5$ iterations for initial exploration. The hyperparameter domains for each classifier were:
\begin{itemize}
\item SVM: type of kernel (linear or radial-basis function kernel), cost $C \in [0.5, 5]$, and gamma coefficient $\gamma \in [0.00001, 1.0]$
\item RF: number of estimators $N_e \in \{5, 10, 15, 20, 25\}$
\item XGB: maximum depth $d \in \{3, 4, 5, 6, 7\}$, learning rate $\lambda \in [0.001, 0.1]$, and number of estimators $N_e \in \{10, 50, 100, 200\}$
\item NB has no hyperparameters
\item kNN: number of neighbors $k \in \{2, 3, 4, 5, 6, 7, 8\}$\end{itemize}
The hyperparameter tuning optimization for the classifiers is also performed in a $K$-Fold Cross Validation setup ($K=5$), but instead of using the whole dataset (as was the case for the SNN) only the training set of the LOOCV partition was used. Similar to the BO for the SNN, the combination of hyperparameters with the best average validation accuracy is chosen for each classifier.

\section{Results}\label{section:results}

\begin{table*}[ht]
    \caption{Comparison between Baseline Features and SNN Extracted Features (among all channels)}
    \label{table:results}
    \centering
    \begin{tabular}{ m{1cm}m{4cm}m{2.5cm}m{2.5cm}m{2.5cm}}
        \toprule
         & Classifier     & Accuracy     & Sensitivity  & Specificity \\
         \midrule
         (i) & FFT-kNN     & $0.60 \pm 0.31$           & $0.56 \pm 0.33$           & $0.64 \pm 0.30$\\
         (ii) & FFT-NB     & $0.57 \pm 0.32$           & $0.33 \pm 0.38$           & $0.85 \pm 0.14$\\
         (iii) & FFT-RF      & $0.58 \pm 0.32$           & $0.58 \pm 0.32$           & $0.64 \pm 0.29$\\
         (iv) & FFT-SVM     & $0.66 \pm 0.28$ & $0.69 \pm 0.26$  & $0.63 \pm 0.29$\\
        (v) & FFT-XGB     & $0.65 \pm 0.28$           & $0.68 \pm 0.26$           & $0.61 \pm 0.30$ \\
        \midrule
        (vi)    & EEGNet \shortcite{eegnet}     & $0.58 \pm 0.32$ & $0.58 \pm 0.31$ & $0.59 \pm 0.32$ \\
        (vii)    & EEGNet-SSVEP \shortcite{eegnet}     & $0.54 \pm 0.34$ & $0.60 \pm 0.31$ & $0.46 \pm 0.37$ \\
        (viii)    & Riemann \shortcite{riemann} & $0.41 \pm 0.50$ & $0.47 \pm 0.54$ & $0.44 \pm 0.50$ \\
        (ix)    & DeepConvNet \shortcite{schirrmeister2017deep} & $0.54 \pm 0.12$ & $0.64 \pm 0.08$ & $0.41 \pm 0.14$ \\
        (x)    & ShallowConvNet \shortcite{schirrmeister2017deep} & $0.57 \pm 0.32$ & $0.58 \pm 0.31$ & $0.56 \pm 0.32$ \\
        \midrule
        (xi)    & DSTFT-SNN-kNN     & $0.78 \pm 0.20$ & $0.78 \pm 0.19$ & $0.77 \pm 0.20$\\
        (xii)   & DSTFT-SNN-NB      & $0.76 \pm 0.21$ & $0.78 \pm 0.20$ & $0.73 \pm 0.23$\\
        (xiii)  & DSTFT-SNN-RF      & $0.79 \pm 0.18$ & $0.81 \pm 0.17$ & $0.77 \pm 0.20$\\
        (ixx)    & DSTFT-SNN-SVM     & $0.78 \pm 0.19$ & $0.83 \pm 0.16$ & $0.72 \pm 0.23$\\
        (xx)     & \textbf{DSTFT-SNN-XGB}     & $\mathbf{0.83 \pm 0.16}$ & $\mathbf{0.84 \pm 0.15}$ & $\mathbf{0.82 \pm 0.16}$\\
        \bottomrule
    \end{tabular}
\end{table*}


Classification results observed, with the extracted features from the proposed SNN, are compared with 
state-of-the-art classifiers developed by \citet{schirrmeister2017deep}, \citet{riemann} and \citet{eegnet}. 
We further compare our approach against classifiers able to learn directly from spectral/FFT features extracted each channel \cite{hindarto2016feature}. The EEG classifiers proposed in previous work are referred to as: (vi) EEGNet, (vii) EEGNet-SSVEP, (viii) Riemann, (ix) DeepConvNet, (x) ShallowConvNet.  The FFT features classifiers are referred to as: (i) FFT-kNN, (ii) FFT-NB, (iii) FFT-RF, (iv) FFT-SVM, (v) FFT-XGB. The proposed classifiers based on the SNN extracted features are referred to as: (xi) DSTFT-SNN-kNN, (xii) DSTFT-SNN-NB, (xiii) DSTFT-SNN-RF, (ixx) DSTFT-SNN-SVM, (xx) DSTFT-SNN-XGB.

According to Table \ref{table:results}, the SNN features outperform the baselines considered by an average of 20pp both in accuracy, specificity and sensitivity. In fact all of the collected differences are statistically significant under significance thresholds below 1E-5. 

The results observed when considering FFT features underline the difficult nature of the problem at hands, showing that the use of spectral features is not sufficient to capture discriminative electrophysiological brain patterns. 

As previously mentioned in Section \ref{section:related_work}, the previous work on EEG -- referred in Table \ref{table:results} as: (vi), (vii), (viii), (ix) and (x) -- is unable to capture neuroplasticity differences between healthy and Schizophrenia individuals from resting state data. These approaches are mainly prepared to detect evoked potentials in response to specific stimuli, thus generally neglecting subtle, spontaneous electrophysiological variations in the brain of individuals.

In contrast, the use of DSTFT representations followed by application of the proposed SNNs are better prepared to detect neuroplasticity characteristics on the EEG signal as motivated by the rich spectral content inputted to the SNN, the properties of the entailed transformations, and the discriminative power of the features outputted from the SNN. These observations are experimentally demonstrated by the results presented in Table \ref{table:results}, with a significant difference between our approach and the previous work on EEG. 

Among the classifiers applied to the SNN features, XGBoost has the better performance, followed by RFs, SVMs with sparse kernel and kNNs. 
We hypothesize that this observation is primarily driven by the 
compositional value of the extracted features and the heterogeneity of individual profiles. Understandably, since only a part of the overall features have discriminative value for a given subject due to profile heterogeneity, NB and kNN have an understandable lower performance due to their inherent inability to discard non-relevant features. 
Similarly, when we compare the classifiers performance from FFT features, FFT-kNN and FFT-NB have a slightly inferior performance against FFT-XGB and FFT-SVM. Among the five classifiers all of them slightly underperformed on discriminating healthy controls (specificity) than discriminating schizophrenic individuals (sensitivity) due to an inherent ability to avoid false negatives. 

The gathered results confirm the relevance of working in a pairwise distance space to guarantee a good generalization ability. In addition, the applied convolution transformations guarantee a sensitivity to the inherently rich spatial, temporal and spectral nature of the EEG signal. We hypothesize that these aspects, together with the use of regularization and the cosine loss function (able to favor variations over absolute differences in the spectral content), explain the ability to learn extremely discriminative features.


\section{Conclusion}\label{section:conclusion}

The rich nature of the electrophysiological data measured at the cerebral cortex makes deep learning a natural candidate to study disorders disrupting the normal brain activity. Nevertheless, the limited size of case-control populations, together with the inherent variability of the spectral content within and among individuals, had left the value of neural network approaches largely unexplored. This manuscript stresses the relevance of revisiting this problem, showing that adequately reshaped neural networks with proper loss and regularization can increase the accuracy of Schizophrenia diagnostics by 15-to-20 percentage points against peer alternatives (without hampering sensitivity or specificity). 

Two master principles underlie these results: 1) the mapping of the original data space into a pairwise distance space to support data augmentation while enhancing the discriminative power of the output features; and 2) the exploration of the rich nature of brain patterning through convolution operations on the spectral imaging of the signal, with weights learned under a cosine loss to improve robustness against the inherent noisy nature of electrophysiologic data. 

As future work, we aim to extend the experimental analysis towards alternative disorders, and different EEG instrumentation or protocols; contrast the performance of the proposed EEG-based learners against state-of-the-art MRI- and PET-based learners on a population of individuals with (and without) neurodegenerative conditions being currently monitored at Instituto de Medicina Molecular; and to establish a method that is capable of performing a neurofeedback technique to tackle Schizophrenia symptoms, similarly to what has been previously proposed by \citet{nan2012individual}.



\bibliographystyle{aaai}
\bibliography{references}
\end{document}